\documentclass[letterpaper, 10pt, conference]{ieeeconf}

\IEEEoverridecommandlockouts
\usepackage{times}
\usepackage{amsmath}
\usepackage{amssymb}
\usepackage{graphicx}
\usepackage{url}
\usepackage{cite}
\usepackage{caption}
\usepackage{algorithm}
\usepackage[noend]{algpseudocode}
\usepackage{booktabs}
\usepackage{xcolor}
\usepackage{subcaption}
\usepackage{tikz}
\usepackage{newfloat}
\usepackage{listings}

\usepackage{booktabs}
\usepackage{balance}

\newcommand{\methodnamelong}{Prismatic World Model}
\newcommand{\methodname}{PRISM-WM}
\makeatletter
\let\origsection\section
\newcommand{\section@nostar}[1]{\origsection{\MakeUppercase{#1}}}
\newcommand{\section@star}[1]{\origsection*{\MakeUppercase{#1}}}
\renewcommand{\section}{\@ifstar{\section@star}{\section@nostar}}
\renewcommand{\thesection}{\Roman{section}}
\renewcommand{\thesubsection}{\Alph{subsection}}
\makeatother
\makeatletter
\renewcommand{\subsubsection}[1]{%
  \par\addvspace{0.8ex}%
  \noindent\textbf{#1}\ \ignorespaces
}
\makeatother
\makeatletter
\renewcommand{\paragraph}[1]{%
  \par\addvspace{0.8ex}%
  \noindent\textbf{#1}\ \ignorespaces
}
\makeatother

\DeclareCaptionStyle{ruled}{labelfont=normalfont,labelsep=colon,strut=off}
\floatstyle{ruled}
\newfloat{listing}{tb}{lst}{}
\floatname{listing}{Listing}

\title{Prismatic World Model: Learning Compositional Dynamics \\ for Planning in Hybrid Systems}

\author{%
Mingwei Li\textsuperscript{1*}, Xiaoyuan Zhang\textsuperscript{2*}, Chengwei Yang\textsuperscript{1\dag}, Zilong Zheng\textsuperscript{3\dag}, Yaodong Yang\textsuperscript{2\dag}%
\thanks{\textsuperscript{*}These authors contributed equally. \textsuperscript{\dag}Corresponding author. Code: \protect\url{https://github.com/SC-Levi/PRISM-WM.git}.}%
\\[-0.1em]
\textsuperscript{1}Beijing Institute of Technology, \textsuperscript{2}Peking University\\[-0.1em]
\textsuperscript{3}NLCo Lab, Beijing Institute for General Artificial Intelligence%
}%

\begin{document}

\maketitle
\vspace{-1.2em}

\begin{abstract}
Model-based planning in robotic domains is challenged by the hybrid nature of physical dynamics, where continuous motion is punctuated by discrete events such as contacts and impacts. Conventional latent world models typically employ monolithic neural networks that enforce global continuity, which over-smooths distinct dynamic modes (e.g., sticking vs. sliding, flight vs. stance). For a planner, this smoothing results in compounding errors during long-horizon lookaheads, rendering the search process unreliable at physical boundaries. To address this, we introduce the Prismatic World Model (PRISM-WM), a structured architecture designed to decompose complex hybrid dynamics into composable primitives. PRISM-WM uses a context-aware Mixture-of-Experts (MoE) framework where a gating mechanism implicitly identifies the current physical mode, and specialized experts predict the associated transition dynamics. We further introduce a latent orthogonalization objective to ensure expert diversity, preventing mode collapse. By modeling the mode transitions in system dynamics, PRISM-WM reduces rollout drift. Experiments on continuous control benchmarks, including high-dimensional humanoids and multi-task settings, demonstrate that PRISM-WM provides a high-fidelity substrate for trajectory optimization algorithms (e.g., TD-MPC), indicating its potential as a foundational model for model-based agents.
\end{abstract}

\section{Introduction}
The efficacy of Model-Based Reinforcement Learning (MBRL) and planning is bounded by the predictive fidelity of the underlying world model \cite{janner2019trust}. While latent variable models enable planning from high-dimensional observations by abstracting away pixel details \cite{ha2018world, hafner2025mastering}, they face challenges in robotic domains due to the hybrid nature of real-world physics. Legged locomotion and manipulation involve systems that continuously switch between discrete contact modes, resulting in non-smooth and highly non-linear dynamics. Specifically, these contact-rich interactions introduce complementarity constraints that generate sharp discontinuities in the underlying vector field.

Prevailing approaches typically approximate complex dynamics using single, monolithic neural networks \cite{hansen2022temporal, georgiev2024pwm}. While theoretically capable of universal approximation, these models practically learn an averaged representation that over-smooths the discontinuities inherent in multi-contact environments. For a trajectory optimization planner, this over-smoothing induces compounding errors at physical boundaries, forcing reliance on restricted lookahead horizons and compromising long-term reasoning. Notably, this spectral blurring across contact boundaries results in significant gradient mismatch, inherently hindering the precise propagation of planning signals during optimization.

To overcome this bottleneck, world models must incorporate structural priors derived from hybrid dynamical systems. PRISM-WM achieves this by learning a data-driven decomposition of specialized latent regimes. Rather than forcing a global function to fit piecewise physics, our context-aware gating network identifies the active physical mode, routing the state to expert networks that capture the continuous flow within that specific regime. Analogous to a prism decomposing polychromatic light into distinct spectra, our orthogonalization layer refines the latent space into independent basis vectors that structurally align with the discrete transitions of piecewise dynamics. This architecture provides the structural decomposition necessary for high-fidelity, long-horizon motion generation without sacrificing the flexibility of deep neural learning.

In this work, we introduce the Prismatic World Model (PRISM-WM). PRISM-WM decomposes the transition function using a context-aware Mixture-of-Experts (MoE) architecture. A gating network routes each context to a weighted combination of experts, implicitly identifying the active physical regime while each expert specializes in local dynamics to avoid interference. We theoretically and empirically show that this decomposition reduces the one-step model residual from $\varepsilon$ to approximately $\varepsilon/K^{\alpha}$, effectively curbing the compounding drift in high-dimensional configuration spaces.

\begin{figure}[t]
    \centering
    \includegraphics[width=0.97\columnwidth]{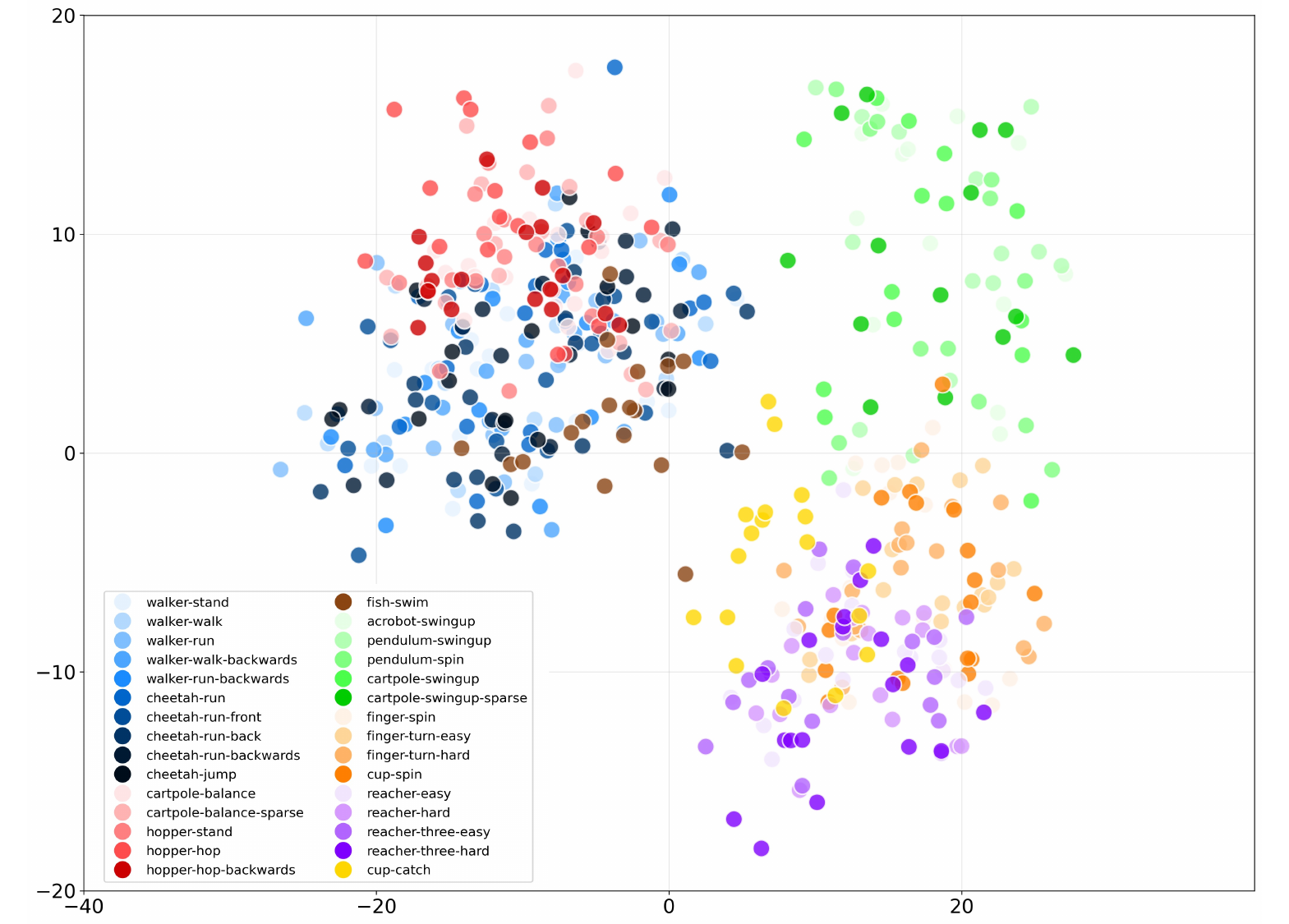}
    \caption{Physical clustering on DMC-30. t-SNE of gating weights across 30 tasks. Without explicit supervision, PRISM-WM clusters tasks by shared physical properties, disentangling high-inertia locomotion from precision manipulation.}
    \label{fig:tsne_performance}
    \vspace{-0.5cm}
\end{figure}

We integrate PRISM-WM into the TD-MPC planning framework \cite{hansen2022temporal}. Experiments across continuous control domains demonstrate that PRISM-WM lowers prediction error over longer horizons compared to monolithic baselines. The model learns to cluster states into physical modes (Fig.~\ref{fig:tsne_performance}) without explicit supervision. By providing a geometry-preserving representation that respects the intrinsic manifold of hybrid systems, PRISM-WM enables the planner to solve tasks that monolithic baselines fail to master.

\section{Related Work}

\subsection{Mixture-of-Experts in RL}
The Mixture-of-Experts (MoE) architecture \cite{jacobs1991adaptive}, foundational to scaling large language models \cite{du2022glam,zoph2022design,clark2022unified,riquelme2021scaling,zhou2022mixture}, has been adapted in reinforcement learning primarily for policies and value functions. Recent works use MoE for multi-task transfer \cite{hendawy2023multi,cheng2023attentionmoe}, probabilistic ensembles \cite{ren2021pmoe}, and hierarchical or continual RL \cite{akrour2021continuous,deramo2020sharing,wang2022expertmix,zhou2021modem}. Closer to our domain, NMOE \cite{karnachan2020nested} explores nested mixtures for identifying hybrid dynamical systems. Distinct from the hierarchical structure of NMOE, PRISM-WM introduces a flat, context-aware routing mechanism coupled with latent orthogonalization. This design models diverse physical dynamics and prevents mode collapse during long-horizon planning.

\subsection{Model-Based Reinforcement Learning and Planning}
Model-Based RL (MBRL) learns predictive models to facilitate data-efficient planning \cite{jimenez2012review, arora2018review, asai2018classical, pasula2007learning}. Recent advances use latent dynamics for synthetic rollouts \cite{heess2015svg, chua2019pets, hafner2019planet, janner2019trust, clavera2020maac, zhang2025differentiable}, value estimation \cite{schrittwieser2020muzero}, and representation learning \cite{konidaris2018skills, zhang2025world}. The Dreamer series \cite{hafner2019dream, hafner2025mastering} trains policies entirely within latent imagination, TD-MPC \cite{hansen2022temporal, hansen2023td} conducts online model-predictive control in latent space, and PWM \cite{georgiev2024pwm} integrates differentiable planning. PRISM-WM is complementary to these downstream control paradigms. By factorizing complex hybrid dynamics into non-redundant experts without manual engineering \cite{fern2011first}, it provides a domain-independent dynamics substrate that mitigates the compounding errors bottlenecking current MBRL algorithms.

\begin{figure}[t!]
    \centering
    \includegraphics[width=1.0\columnwidth, trim={12mm 0.0cm 30mm 0cm}, clip]{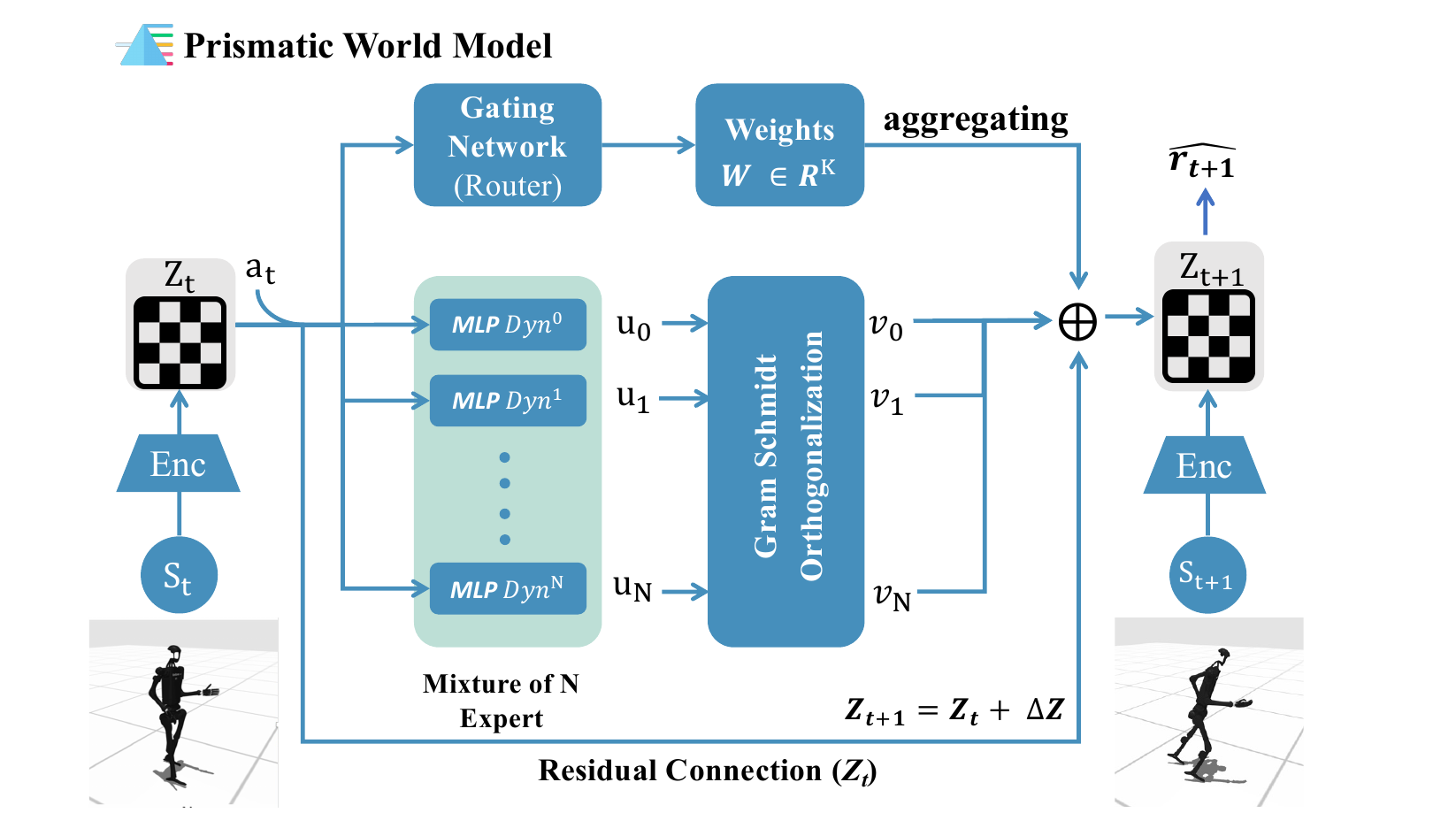}
    \caption{The PRISM-WM architecture. To capture hybrid dynamics, the model decomposes transitions: the gating network identifies the active latent regime, while orthogonal experts learn a non-redundant basis for the residual dynamics $\Delta Z$, preventing mode collapse during planning.}
    \label{fig:prism_arch}
\end{figure}

\section{Method}
\label{sec:method}

Our work introduces the Prismatic World Model (PRISM-WM), a world model architecture designed to enhance online planning and direct policy learning paradigms within MBRL. We first review the fundamentals of latent world models and the two mainstream frameworks they support, highlighting their shared dependency on the model's representational fidelity. Then, we detail the architecture of PRISM-WM and explain how it serves as a unified foundation for improving both frameworks.

\subsection{Preliminaries}
\label{sec:preliminaries}

A latent world model predicts future states and rewards within a compact, learned representation space. It typically consists of three core components: an encoder $q(z_t|o_t)$ mapping observations to latent states, a forward dynamics model $p(z_{t+1}|z_t, a_t)$, and a reward model $p(r_t|z_t, a_t)$.  Once trained, the world model acts as a surrogate environment for downstream control, conventionally utilized through online planning (e.g., the TD-MPC framework) or direct policy learning (e.g., the PWM framework).

Both control paradigms depend on long-horizon model fidelity but expose vulnerabilities in standard architectures. In online planning, where algorithms evaluate simulated trajectories to optimize action sequences, the bottleneck is compounding error. Small, single-step prediction inaccuracies in the dynamics model accumulate over the planning horizon, leading to suboptimal decision-making. Direct policy learning uses the world model as a differentiable simulator to update a policy via analytic gradients backpropagated through time. In this setting, the obstacle is gradient quality; a non-smooth or biased dynamic loss landscape yields noisy gradients that destabilize policy optimization. Overcoming these intertwined challenges motivates the architectural design of PRISM-WM.

\begin{algorithm}[t]
\caption{Prismatic World Model}
\label{alg:prism_forward}
\begin{algorithmic}[1]
\State \textbf{Input:} policy $\pi_\theta$, world model $M_\varphi$, 
      replay buffer $\mathcal{D}_{\text{env}}$
\For{$k = 1$ to $K$}                        
    \State Sample trajectories $(s_t,a_t,r_t,s_{t+1})$
           with $\pi_\theta$ add to $\mathcal{D}_{\text{env}}$
    \State Train $M_\varphi$ on $\mathcal{D}_{\text{env}}$ to minimize $J_M(\varphi)$

    \For{$t = 0$ to $T-1$}                  
        \State Sample $(z_t,a_t,r_t)$
               via interaction of $\pi_\theta$ \& $M_\varphi$

        \State $x_t \gets [\,z_t, a_t\,]$
        \State $\mathbf{w} \gets \operatorname{softmax}(g_{\text{gate}}(x_t)/\tau_t)$

        \State $u_k \gets E_k(x_t)$ for $k=1\dots K$

        \For{$k = 1$ to $K$}
            \State $\tilde{v}_k \gets u_k - \sum_{i=1}^{k-1} \langle v_i, u_k \rangle \, v_i$
            
            \State $v_k \gets \frac{\tilde{v}_k}{\|\tilde{v}_k\|_2 + \epsilon}$ 
        \EndFor

        \State $\displaystyle
               \hat z_{t+1}
               = z_t + \sum_{k=1}^{K} w_{t,k}\,v_k$   

        \For{$j = 0$ to $H-1$}
            \State $a_j \gets \pi_\theta(z_j)$
            \State $z_{j+1} \gets M_\varphi(z_j,a_j)$
        \EndFor

        \State $\displaystyle
               \mathcal{L}_\pi =
               -\frac{1}{H}\sum_{j=0}^{H-1}
               \Bigl[\,Q_\psi(z_j,a_j)
               + \lambda\,\mathcal{H}(\pi_\theta(\cdot|z_j))\Bigr]$
        \State $\theta \gets \theta - \alpha\nabla_\theta \mathcal{L}_\pi$

    \EndFor
\EndFor
\State \Return Optimal policy $\pi_\theta$
\end{algorithmic}
\end{algorithm}

\begin{figure}[t!]
    \centering
    \includegraphics[width=1.0\columnwidth, trim={0cm 0.5cm 10mm 1.0cm}, clip]{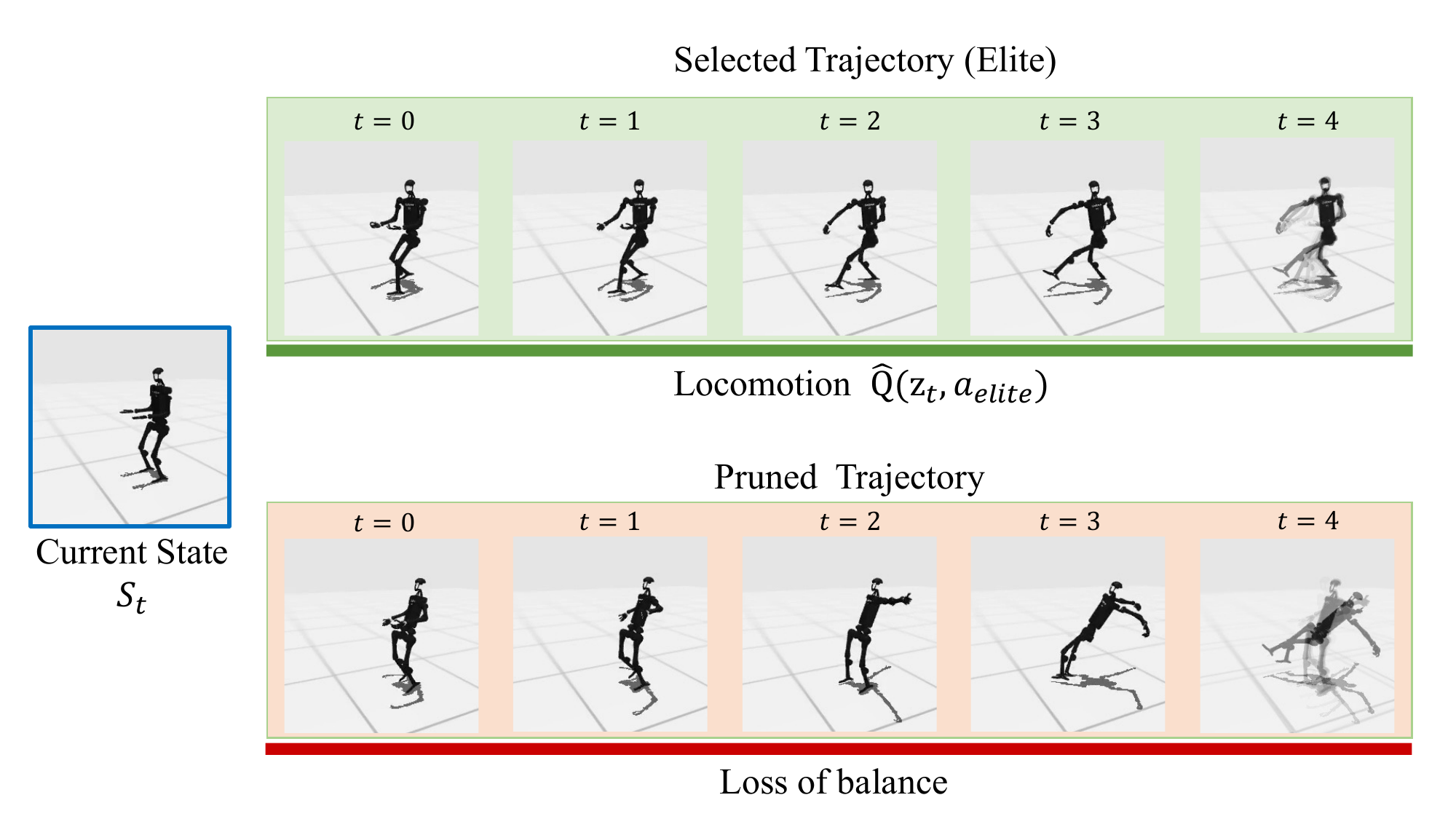}
    \caption{High-fidelity lookahead prevents physical violations. (Top) A stable trajectory (green) maintaining locomotion. (Bottom) A pruned branch predicting a loss of balance (red). \textbf{By forecasting physical boundaries, the planner rejects unsafe actions before execution.}}
    \label{fig:prism_unrolled}
    \vspace{-0.3cm}    
\end{figure}

\subsection{Architecture Details}
\label{sec:architecture}
PRISM-WM employs a set of $K$ expert networks $\{E_k\}_{k=1}^K$ and a single gating network $G$. For predicting the latent state transition, the gating network receives a context vector $c_t$ and produces softmax-normalized weights $w_t \in \mathbb{R}^K$. The next state $\hat{z}_{t+1}$ is computed as a weighted combination of expert outputs with a residual connection to the current state $z_t$:
\begin{equation}
\label{eq:prism_dynamics}
\hat{z}_{t+1} = z_t + \sum_{k=1}^K w_{t,k} \cdot v_k(x_t),
\end{equation}
where $w_t = \text{softmax}(G_{\phi}(c_t)/\tau_t)$ and $x_t$ is the input to the experts. The forward pass is detailed in Algorithm~\ref{alg:prism_forward}. 

The design of $c_t$ determines the decomposition. In single-task settings, we define $c_t = [z_t, a_t]$, while multi-task settings yield $c_t = [z_t, a_t, e_{\text{task}}]$. This allows the model to capture task-specific dynamic variations while retaining the ability to handle multi-modal dynamics within each task. 

To prevent the experts from learning redundant features, we introduce an orthogonalization layer inspired by MOORE \cite{hendawy2023multi}. After each expert $E_k$ outputs a penultimate-layer vector $u_k(s) \in \mathbb{R}^d$, we apply the differentiable Gram-Schmidt process to the expert stack $U_s = [u_1(s), \dots, u_K(s)] \in \mathbb{R}^{d \times K}$. Ideally, $U_s$ lies on the Stiefel manifold:
\begin{equation}
U_s^{\top} U_s = I_K, \qquad \forall s \in \mathcal{S}. \tag{4}
\end{equation}
Instead of explicit constraints, we apply a stabilized, differentiable Gram-Schmidt operator to the expert outputs $u_k = E_k(x_t)$. For each $k \in \{1, \dots, K\}$, we first compute the orthogonal residual $\tilde{v}_k$, followed by its normalization:
\begin{equation}
\label{eq:stabilized_gs}
\tilde{v}_k = u_k - \sum_{i=1}^{k-1} \langle v_i, u_k \rangle v_i, \quad v_k = \frac{\tilde{v}_k}{\|\tilde{v}_k\|_2 + \epsilon}, \tag{5}
\end{equation}
where $\epsilon$ is a small constant ensuring the Jacobian remains well-conditioned during backpropagation.

While the Gram-Schmidt operator imposes a sequential inductive bias on the basis vectors, physical contact modes (e.g., left versus right foot stance) are inherently unordered and symmetric. Consequently, the individual orthogonal experts $v_k$ do not map one-to-one to specific physical events. Rather, the orthogonalization ensures the experts span a maximal-rank, non-redundant subspace for the residual dynamics. The context-aware gating network $G$ resolves this hierarchy by dynamically routing the state to a weighted combination of these bases, reconstructing the unordered physical modes. By isolating conflicting physical modes, PRISM-WM mitigates compounding drift and provides the planner with distinct trajectories.

\subsection{Mathematical Formalization and Stability Analysis}
\label{sec:stability}

Let $\Phi(x_t) \in \mathbb{R}^d$ denote a shared encoding that extracts task-relevant features from state-action pairs. The orthogonalization layer operates on expert representations $u_k = E_k(\Phi(x_t))$, enforcing the basis set $\{\mathcal{V}_k\}$ as a structural prior. PRISM-WM models the dynamics as a sum of basis transitions:
\begin{equation}
\label{eq:projected_dynamics}
\Delta z_t = \sum_{k=1}^K w_{t,k} \cdot \text{proj}_{\mathcal{V}_k} (\Phi(x_t)),
\end{equation}
where $\text{proj}_{\mathcal{V}_k}$ ensures each expert captures non-redundant components of the residual manifold.

To bridge the gap between latent geometric constraints and physical reality, we evaluate the Mutual Information (MI) between routing weights $w_t$ and ground-truth boolean contact tensors from the MuJoCo simulator. High MI alignment confirms that enforcing a non-redundant basis prevents representation collapse and encourages the routing mechanism to align with distinct physical boundaries. For numerical implementation, we employ a stabilized Gram-Schmidt process with $\epsilon$-perturbation. This ensures a well-conditioned Jacobian, preventing gradient explosion while maintaining the transition as a stable convex combination of Lipschitz experts.

\subsection{Framework for Planning and Policy Learning}
\label{sec:unifying}
PRISM-WM translates the structural orthogonality formulated in Sec.~\ref{sec:stability} into tangible operational advantages for dominant MBRL paradigms. The Stiefel manifold constraint directly facilitates partitioning the non-smooth manifold into $K$ low-complexity physical regimes. For planning-centric methods like TD-MPC2, this reduces the one-step model residual from $\varepsilon$ to roughly $\varepsilon/K^{\alpha}$, curbing the compounding drift inherent in long-horizon rollouts. Furthermore, for policy-centric MBRL like PWM \cite{georgiev2024pwm}, the orthogonal basis maintains a well-conditioned Jacobian, ensuring stable gradient propagation across discontinuous boundaries. This transforms fractured loss landscapes into functionally smooth convex blends, providing high-fidelity signals that enable stable convergence in high-dimensional configuration spaces.

\begin{figure*}[t]
    \centering
    \includegraphics[width=0.97\textwidth]{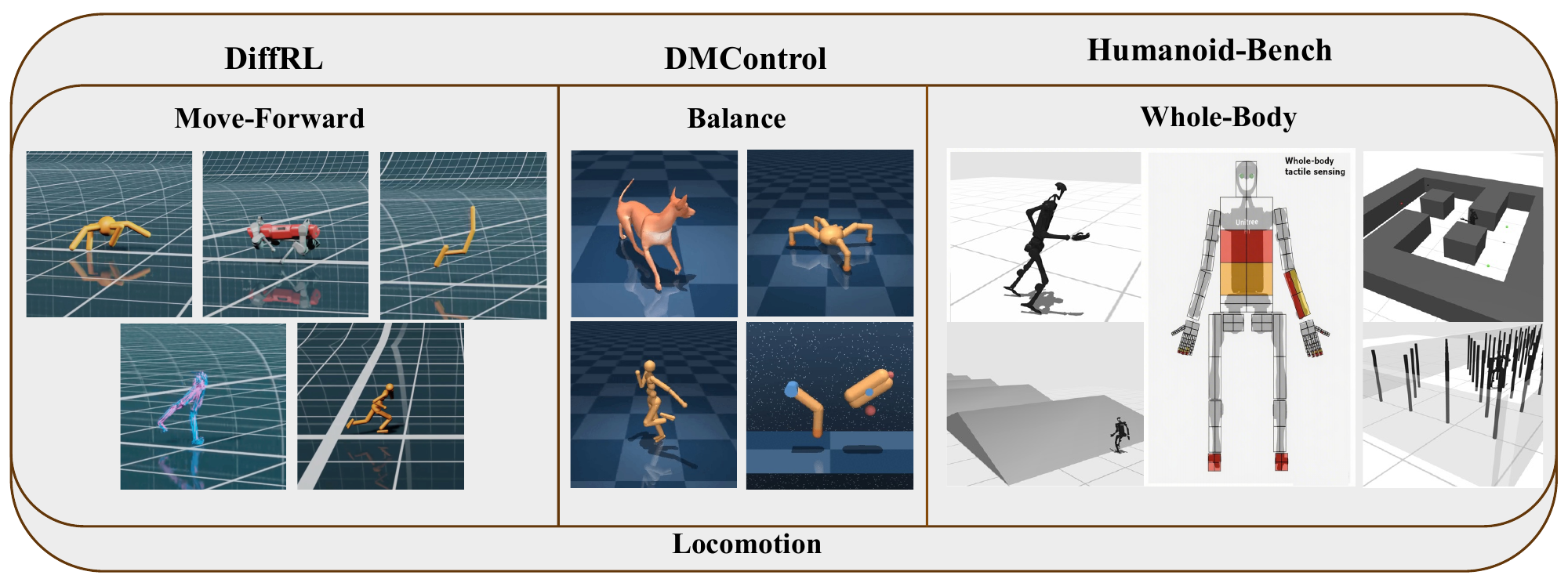} 
    \caption{Performance on high-dimensional locomotion tasks. The plots show the mean episode reward versus environment steps averaged over 5 random seeds (shaded regions represent one standard deviation). PRISM-WM achieves higher sample efficiency and asymptotic performance compared to baselines on high-dimensional humanoid tasks.}
    \label{fig:task_gallery}
    \vspace{-0.3cm}    
\end{figure*}

\begin{figure*}[t]
    \centering
    \includegraphics[width=1.0\textwidth]{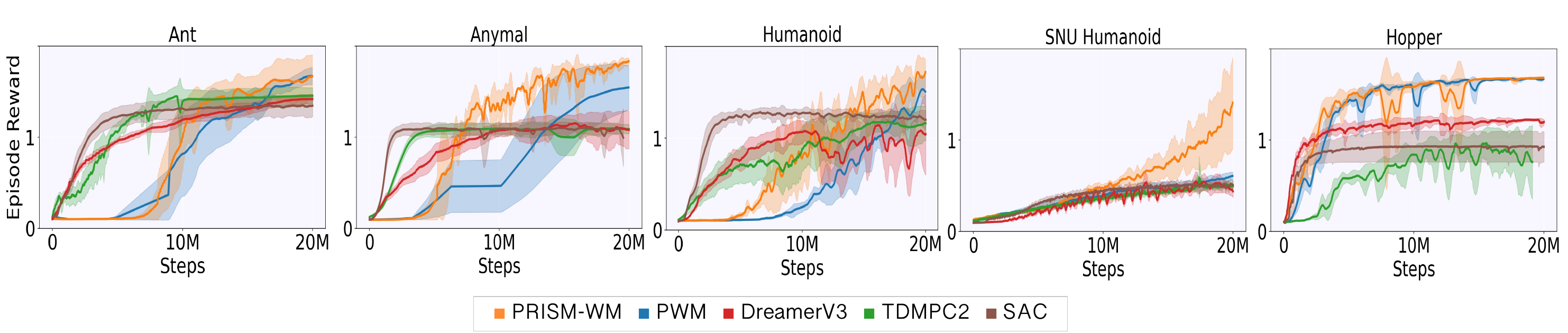}
    \caption{Sample efficiency and policy stability on DiffRL. Mean episode reward vs. environment steps (5 random seeds; shaded regions denote $\pm 1$ SD). \textbf{\methodname{} prevents the policy collapse observed in monolithic models on contact-rich domains (e.g., SNU-Humanoid), leading to higher asymptotic performance.}}
    \label{fig:performance_curves}
    \vspace{-0.3cm}    
\end{figure*}

\section{Experiments}
\label{sec:experiments}




We systematically benchmark \methodname{} against state-of-the-art model-based and model-free methods on continuous control tasks, assessing its efficacy as a structural enhancement for online planning and direct policy learning, alongside its physical robustness in zero-shot sim-to-sim transfer.

We select continuous control environments spanning multiple simulators, including standard locomotion tasks from DMControl \cite{tassa2018deepmind}, whole-body control tasks from Humanoid-Bench \cite{sferrazza2024humanoidbench}, and high-dimensional contact-rich environments from DiffRL \cite{xu2022accelerated}. This rigorously evaluates world model generalization across domains with heterogeneous dynamics.

\methodname{} is integrated into its two target paradigms, termed PRISM-WM. For consistency we retain TD-MPC2 and PWM hyperparameters, replacing only monolithic dynamics and reward models with the MoE-based architecture. We fix the number of experts to $K=4$ for primary evaluations to balance representational capacity and computational efficiency. Comprehensive ablation studies isolate the effects of planning horizon, expert cardinality, and inference efficiency. Finally, visualizations elucidate the interpretable mode-switching behavior of the gating mechanism.

\subsection{High-Dimensional Locomotion on DiffRL}

We evaluate continuous control on five proprioceptive locomotion tasks from the DiffRL suite, with episodes truncated at 1,000 steps based on physical failure criteria. PRISM-WM is integrated into the TD-MPC2 planning framework with original hyperparameters identical. We benchmark against monolithic TD-MPC2 \cite{hansen2023td}, PWM \cite{georgiev2024pwm}, DreamerV3 \cite{hafner2025mastering}, and Soft Actor-Critic (SAC) \cite{haarnoja2018soft}, aggregating results over 5 random seeds.

As depicted in Fig.~\ref{fig:performance_curves}, \methodname{} achieves higher sample efficiency and final returns across the evaluated domains compared to the baselines. On the quadrupedal tasks (Ant, Anymal) and Hopper, the architecture converges more rapidly. In the high-dimensional environments (Humanoid and SNU-Humanoid), the monolithic TD-MPC2 and model-free baselines experience policy collapse. This failure stems from compounding prediction errors when monolithic models average the non-smooth contact events of bipedal locomotion. By decomposing these dynamics into a composable basis, PRISM-WM reduces physical boundary errors. This dynamics substrate prevents the planner from exploiting false simulator states, enabling stable optimization where standard world models fail.

\begin{figure}[t]
    \centering
    \includegraphics[width=0.97\columnwidth]{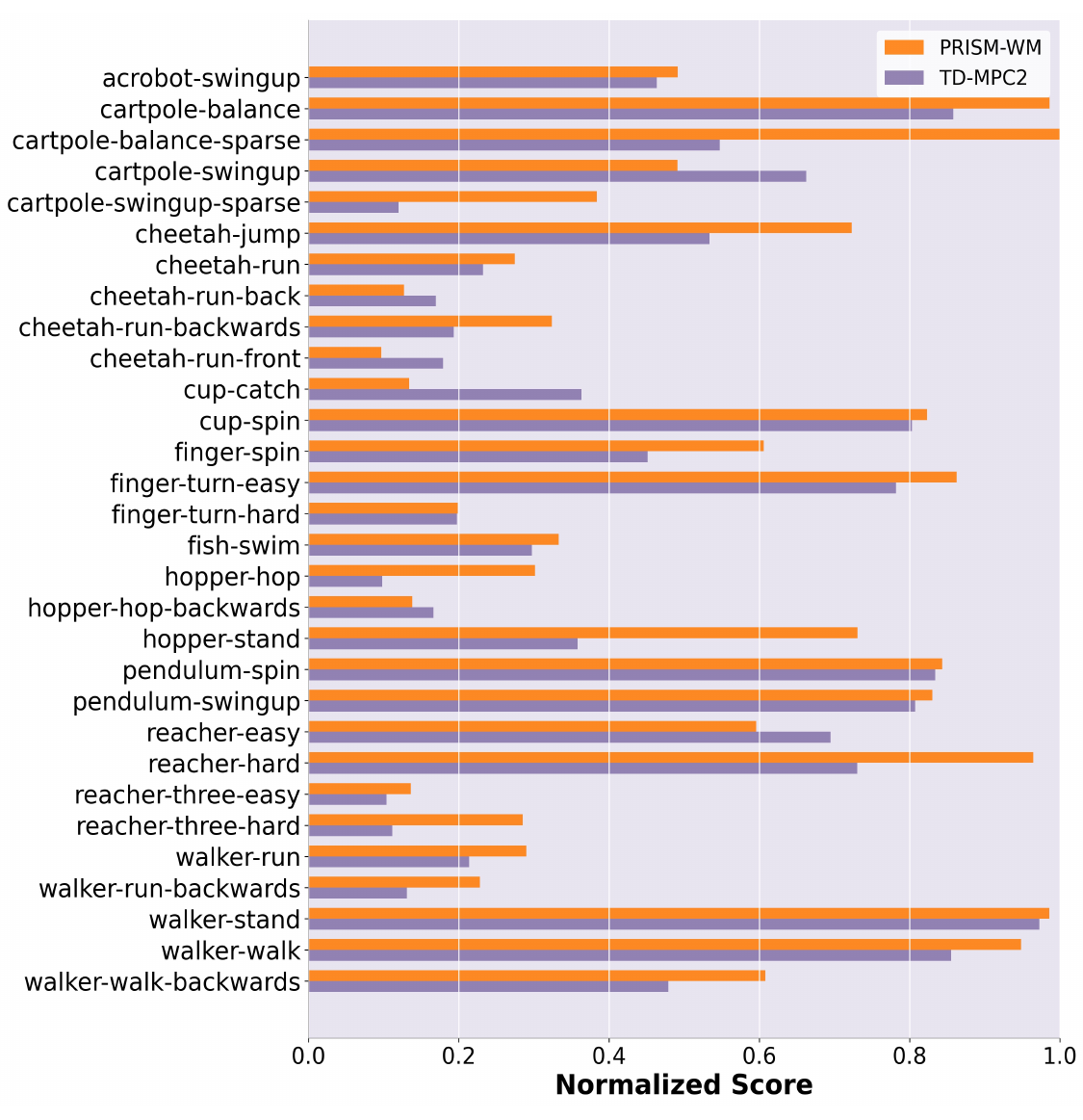}
    \caption{Robust multi-task generalization across 30 domains. PRISM-WM achieves the highest mean normalized score of 0.531. \textbf{Our orthogonalized MoE architecture effectively mitigates task interference}, significantly outperforming the monolithic TD-MPC2 (0.430) across a diverse spectrum of physical environments.}
    \label{fig:mt30_performance}
\end{figure}

\subsection{Multi-Task Generalization}

To rigorously evaluate PRISM-WM in mitigating negative transfer across heterogeneous dynamics, we train a single agent simultaneously on a custom 30-task DeepMind Control Suite benchmark. We report the mean normalized score after 3 million environment steps, comparing PRISM-WM against the monolithic TD-MPC2 baseline.

As illustrated in Fig.~\ref{fig:mt30_performance}, PRISM-WM attains a mean normalized score of 0.531, yielding a 23.5$\%$ improvement over TD-MPC2 (0.430). It surpasses the baseline on 26 of 30 tasks, averaging a +0.12 per-task gain. Consequently, the mixture-of-experts architecture prevents mode collapse within high-variance locomotion domains. PRISM-WM registers score increases of +0.31 on cheetah-run-backwards and +0.28 on reacher-three-hard, where the baseline struggles with severe gradient interference during joint optimization. Furthermore, performance on solved manipulation tasks (e.g., finger-spin) remains stable ($\pm 0.02$), indicating the gating mechanism improves complex domain performance without deteriorating established skills.

We attribute this robust generalization to the gating network's ability to cluster tasks with similar physical properties (e.g., Locomotion vs. Balance) into distinct expert active regions. By partitioning conflicting global dynamics into specialized sub-models, PRISM-WM explicitly isolates task-specific knowledge, thereby establishing a scalable generalist substrate for multi-skill continuous control.

\subsection{High-Dimensional Humanoid Control}

We evaluate PRISM-WM in high-dimensional state spaces with unstable hybrid dynamics using the Humanoid-Bench suite. Control in these domains requires accurate predictions of continuous locomotion interspersed with contact discontinuities, testing a world model's representational fidelity.

\begin{figure}[t]
    \centering
    \includegraphics[width=1.0\columnwidth, trim={12mm 1mm 24mm 0mm}, clip]{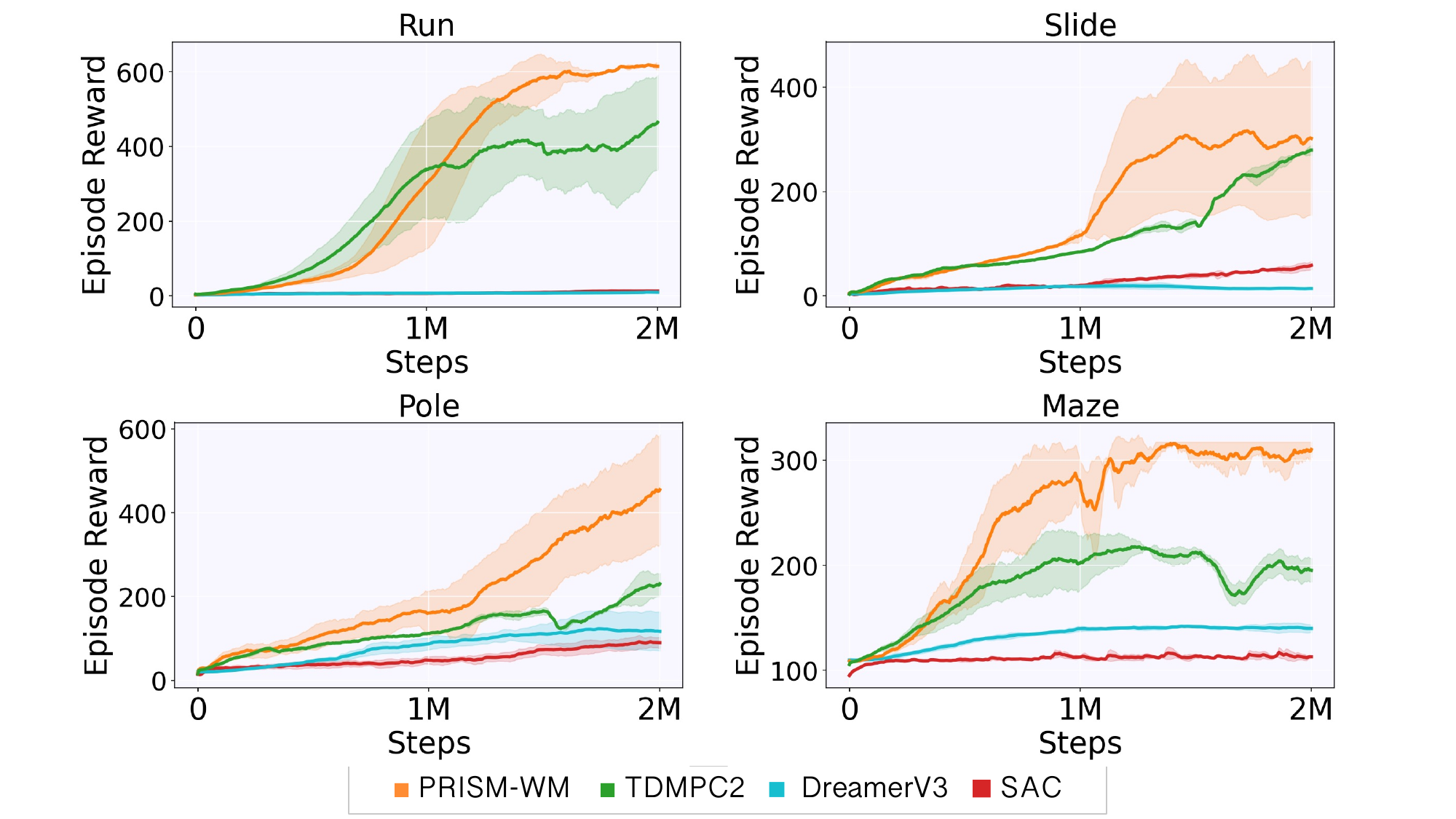}
    \caption{Performance on humanoid control tasks. PRISM-WM attains higher sample efficiency and asymptotic performance than the baselines across four environments, indicating its capacity to model high-dimensional dynamics.}
    \label{fig:humanoid_performance}
\end{figure}

PRISM-WM outperforms the baselines across the four evaluated tasks (Run, Slide, Pole, and Maze) shown in Fig.~\ref{fig:humanoid_performance}. In the Run and Slide tasks, PRISM-WM finds stable gaits and maintains reward growth. The monolithic TD-MPC2 baseline experiences learning volatility and policy collapses, indicating an inability to predict long-horizon contact outcomes. This difference is amplified in the Pole and Maze tasks, where the agent must sustain locomotion while managing interactive objectives such as balancing a pole or avoiding boundaries. The performance of PRISM-WM indicates that the MoE architecture prevents the interference between objectives that degrades monolithic models. By structurally isolating these dynamic modes, PRISM-WM establishes a foundation for optimizing whole-body behaviors.

\begin{table}[t]
    \centering
    \caption{Zero-Shot Sim-to-Sim Transfer Performance}
    \label{tab:sim2sim}
    \setlength{\tabcolsep}{4pt} 
    \footnotesize 
    \begin{tabular}{@{} l c c c c @{}}
    \toprule
    \textbf{Model} & \textbf{Source Return} & \textbf{Target Return} & {\textbf{Drop}} & \textbf{Succ.} \\
    & \text{(IsaacGym)} & \text{(MuJoCo)} & {\text{($\Delta$, \%)}} & \text{(\%)} \\
    \midrule
    MLP (Baseline)  & $297.0 \pm 10$ & $155.0 \pm 38$ & 47.8 & 12 \\
    MoE (Vanilla)   & $402.0 \pm 12$ & $242.0 \pm 25$ & 39.8 & 24 \\
    \textbf{PRISM-WM} & \boldmath{$438.0 \pm 8$} & \boldmath{$386.3 \pm 15$} & \textbf{11.8} & \textbf{88} \\
    \bottomrule
    \end{tabular}
\end{table}

\begin{figure*}[t] 
    \centering
    \includegraphics[width=0.95\textwidth]{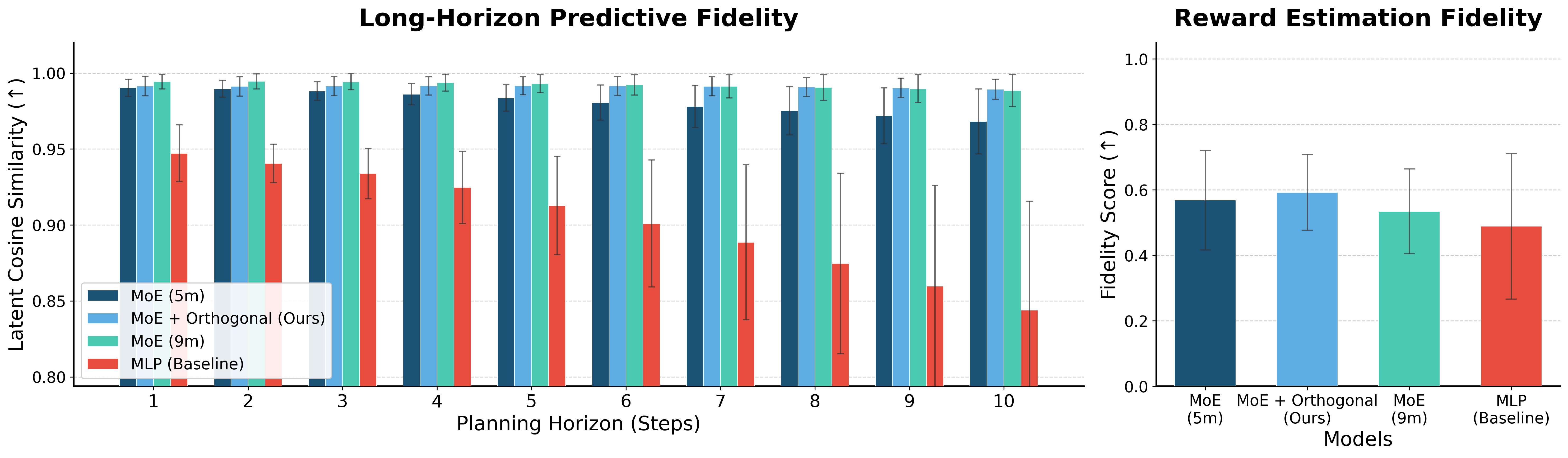} 
    \caption{Structural constraints prevent compounding drift. (Left) Latent cosine similarity. As the prediction horizon extends, monolithic MLP experiences structural collapse, whereas \textbf{orthogonalized MoE maintains manifold fidelity}. (Right) Reward estimation fidelity. Without orthogonal constraints, over-parameterized experts (9M) learn \textbf{redundant representations}, degrading reward estimation compared to the \textbf{structurally constrained model} (5M). Error bars indicate 95\% CI.}
    \label{fig:prediction_error}
\end{figure*}

\begin{figure}[t]
    \centering
    \includegraphics[width=0.97\columnwidth]{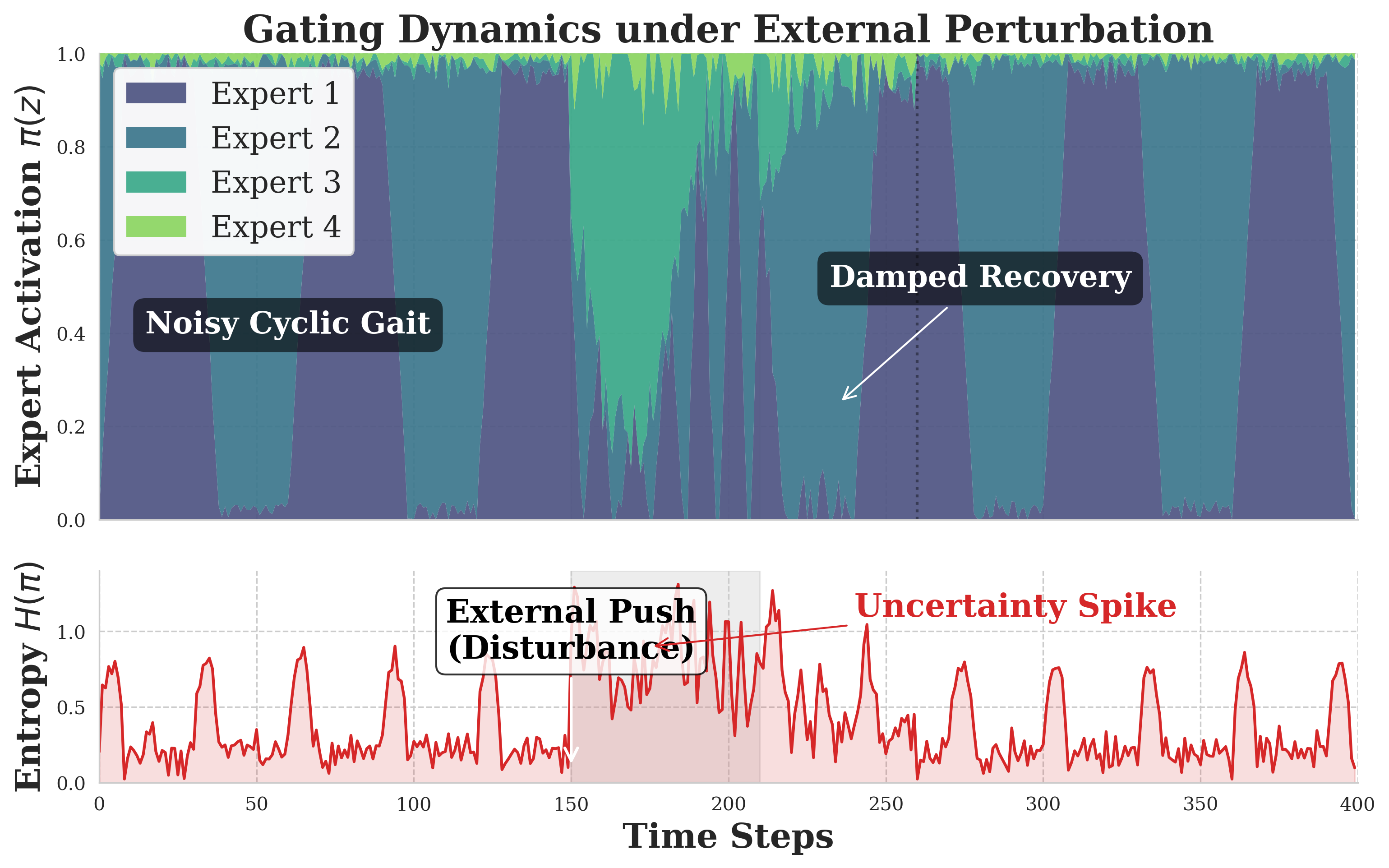}
    \caption{Discrete switching under perturbation. The model maintains \textbf{low entropy} during steady-state locomotion and \textbf{shifts modes} upon physical impact, recovering \textbf{quasi-discrete hybrid dynamics} without the over-smoothing typical of monolithic baselines.}
    \label{fig:gating}
\end{figure}

\begin{figure}[t]
    \centering
    \includegraphics[width=0.9\columnwidth]{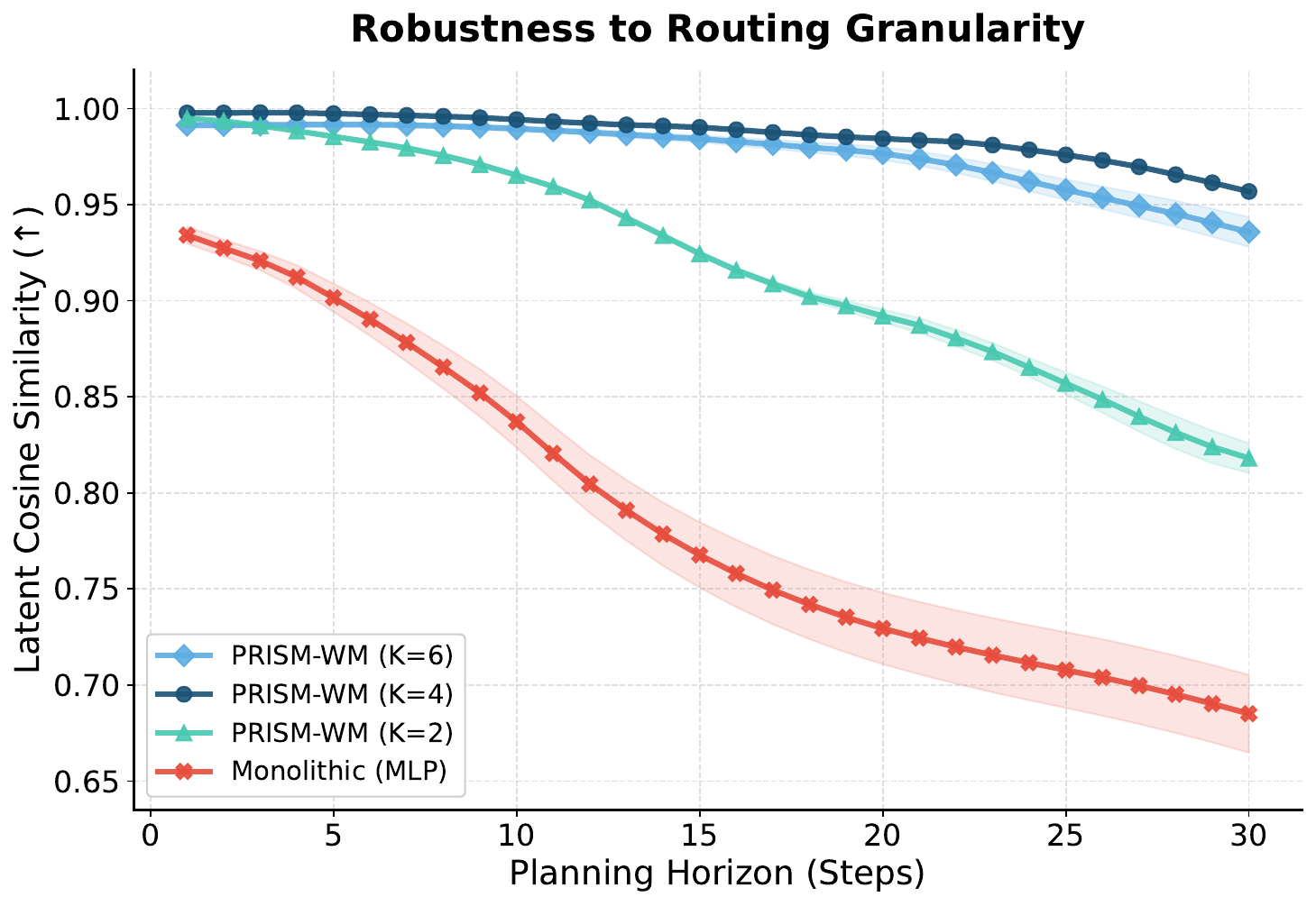}
    \caption{Orthogonal decomposition ensures stability. Latent predictive fidelity evaluated up to $H=30$ across varying expert counts ($K$). As the rollout horizon lengthens, \textbf{standard} monolithic baseline experiences compounding drift, whereas \textbf{PRISM-WM variants maintain stability} \textbf{by preserving latent structure}. The minimal intra-MoE variance indicates that performance gains stem from \textbf{structural decomposition} rather than hyperparameter tuning.}
    \label{fig:ablation_horizon} 
\end{figure}

\subsection{Zero-Shot Sim-to-Sim Transfer}
To evaluate physical grounding, we conduct zero-shot transfer from IsaacGym (soft-contact) to high-fidelity MuJoCo (stiff-contact) across four \textit{humanoid-bench} tasks: \textit{Run, Maze, Pole,} and \textit{Slide}. These environments expose a fundamental bottleneck in high-dimensional MBRL: the coordination of agile upper-body maneuvers with rigid lower-body contact constraints. As summarized in Table \ref{tab:sim2sim}, the monolithic MLP baseline suffers a catastrophic $47.8 \pm 5.2\%$ performance collapse with a $12\%$ success rate. This failure stems from over-smoothing contact boundaries, leading to erroneous foot-slip predictions that destabilize the agent during solver transitions. Crucially, while a vanilla MoE provides marginal improvements, it remains prone to mode interference. In stark contrast, PRISM-WM maintains a 386.3 target return with only an 11.8$\%$ drop, achieving a consistent $88\%$ success rate across both source and target domains. Notably, orthogonalization constrains vertical support forces and horizontal friction residuals to distinct basis subspaces. Consequently, when the reality gap alters friction coefficients, the gating network only needs to adjust the weights of horizontal experts without compromising the structural integrity of the global dynamics manifold.

As visualized in Fig.~\ref{fig:gating}, under a $60\mathrm{N}$ lateral perturbation, the gate exhibits a sharp phase shift toward specialized stabilization experts. Quantitatively, the gating entropy $H(w)$ spikes to $0.8$ during the impulse, reflecting a deliberate transition from rhythmic gait to recovery modes. While monolithic baselines suffer from sustained oscillations, PRISM-WM exhibits a significantly accelerated decay of the limit cycle perturbation, executing a damped return to nominal gait within $1.2\mathrm{s}$. This emergent switching confirms that PRISM-WM captures a transferable and physically consistent decomposition of hybrid robotic physics.

\subsection{Ablation Study}
\label{sec:ablation}
A core hypothesis is that structural decomposition mitigates compounding errors in long-horizon planning. We evaluate this by analyzing rollout fidelity over extended horizons and its sensitivity to the number of experts $K$.

\paragraph{Fidelity of Long-Horizon Value Estimation}
\newline
We evaluate predictive fidelity over extended horizons (Fig.~\ref{fig:prediction_error}).
We utilize Latent Cosine Similarity to measure the directional preservation of state trajectories, alongside a bounded exponential kernel to quantify reward estimation fidelity. While the scaled vanilla MoE matches the dynamics fidelity of our method, it exhibits high predictive variance and degraded reward accuracy (Fig.~\ref{fig:prediction_error} Right). Furthermore, unconstrained over-parameterization in baselines exacerbates expert redundancy and overfitting to local contact noise. Notably, our orthogonalized model achieves superior reward fidelity with fewer parameters by enforcing non-redundant inductive priors. This structural precision prevents the rapid manifold collapse and physically infeasible states, such as ground penetration, typical of monolithic rollouts. Consequently, by maintaining consistent state trajectories, PRISM-WM preserves the validity of long-term value accumulation.

\paragraph{Robustness to Expert Count ($K$)}
\newline
To evaluate sensitivity to routing granularity, we ablate PRISM-WM across varying expert counts. Fig.~\ref{fig:ablation_horizon} shows all MoE variants outperform the monolithic baseline. Even coarse under-segmentation isolates primary physical dichotomies. We identify $K=4$ as optimal, aligning expert capacity with intrinsic dynamic modes, whereas over-segmentation introduces slight variance via expert redundancy. Intra-MoE performance variance is negligible compared to the gap against the baseline. PRISM-WM's efficacy stems from the principle of structural decomposition.

\paragraph{Real-time Inference Efficiency}
\newline
We evaluated whether the proposed MoE architecture introduces any significant latency overheads that hinder real-time robotic control. We benchmarked the inference performance on a single NVIDIA RTX 4090 GPU. As shown in Table~\ref{tab:inference_speed}, the $K=4$ configuration achieves a throughput of 7,712 Hz, which is faster than the monolithic MLP baseline. Although the total parameter count increases, only a subset of experts is activated per step, optimizing overall effective computation. The $K=2$ configuration consumes minimal GPU memory, offering a lightweight solution. With $K=6$, the throughput remains higher than the baseline.

\begin{table}[t]
    \centering
    \caption{Inference efficiency analysis. Planning latency, throughput, and GPU memory usage evaluated on an NVIDIA RTX 4090. PRISM-WM ($K=4$) achieves \textbf{higher throughput} than the monolithic baseline with \textbf{minimal memory overhead}.}
    \label{tab:inference_speed}
    

    \resizebox{\columnwidth}{!}{
        \begin{tabular}{lcccc}
            \toprule
            \textbf{Model} & \textbf{Inference} & \textbf{Throughput} & \textbf{GPU Mem} & \textbf{Rel.} \\
            & \textbf{(ms)} $\downarrow$ & \textbf{(FPS)} $\uparrow$ & \textbf{(GB)} $\downarrow$ & \textbf{Speed} \\
            \midrule
            MLP (Baseline) & 0.13 & 7499 & 0.09 & 1.00$\times$ \\
            \midrule
            PRISM ($K=2$) & 0.14 & 6983 & \textbf{0.07} & 0.93$\times$ \\
            PRISM ($K=4$) & \textbf{0.13} & \textbf{7712} & 0.11 & \textbf{1.03$\times$} \\
            PRISM ($K=6$) & 0.13 & 7581 & 0.19 & 1.01$\times$ \\
            \bottomrule
        \end{tabular}
    }
    \vspace{-0.2cm}         
\end{table}

\begin{figure}[t!] 
    \centering
    \includegraphics[width=0.97\columnwidth]{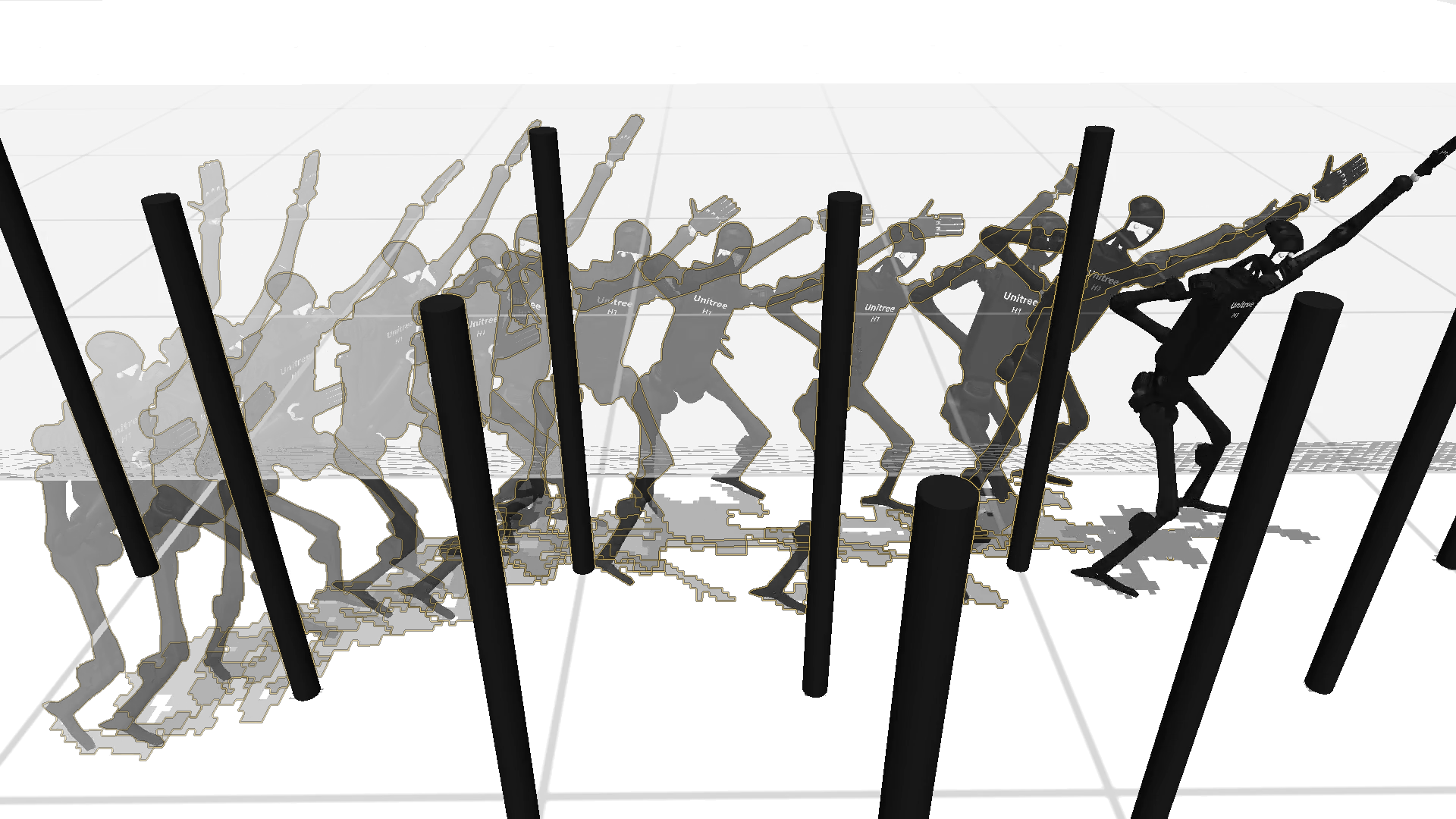}
    \caption{Stroboscopic visualization of whole-body control. Deployed zero-shot in the MuJoCo Pole task, PRISM-WM synthesizes \textbf{torso-dodging maneuvers} while maintaining \textbf{foot-ground contacts}. This demonstrates the \textbf{structural fidelity} required to prevent falls typical of monolithic models during obstacle avoidance.}
    \label{fig:mujoco_pole_vis}
\end{figure}



\subsection{Qualitative Visualization: Agile Whole-Body Control}
\label{sec:qualitative}

While the supplementary video shows all tasks,  the \textit{Pole} task highlights a specific stress test for MBRL. This environment requires the humanoid to weave around obstacles while maintaining stable locomotion, necessitating the coordination of high-frequency, non-smooth upper-body evasive maneuvers with rigid lower-body contact constraints.  

Monolithic models typically fail in this complex regime due to latent manifold collapse when approximating simultaneous free space dodging and stiff ground reactions. As shown in Fig.~\ref{fig:mujoco_pole_vis}, PRISM-WM's orthogonal decomposition enables sharp torso dodging transitions without destabilizing established foot ground contacts. This structural prior ensures robust stability across hybrid boundaries by effectively isolating conflicting dynamic modes into distinct subspaces.




\section{Conclusion}
\label{sec:conclusion}
We introduced \methodnamelong{} (\methodname{}) to address the over-smoothing of hybrid dynamics in monolithic models. \methodname{} outperforms vanilla MoE baselines by enforcing latent orthogonalization, compelling experts to specialize in mutually exclusive dynamics. The gating mechanism thus captures semantic physical regimes, exhibiting distinct mode transitions under perturbations. This structural disentanglement significantly enhances multi-modal representation while mitigating negative interference in complex multi-task settings. Functioning as a versatile backbone for MBRL, \methodname{} ensures robust decision-making over extended planning horizons and stabilizes gradient propagation for policy learning. Decomposing complex physics into orthogonal primitives remains fundamentally essential for mastering contact-rich control. Our results demonstrate \methodname{} mitigates long-term drift, establishing a physically-grounded substrate for future general-purpose embodied agents.

\section{Limitations and Future Work}
\label{sec:Limitations and Future Work}
The current framework has limitations. The expert count $K$ remains a manually tuned hyperparameter; automating its selection is a necessary extension. Exploring composition functions beyond a weighted sum may also improve performance. Applying the model to real-world robotics will test if the orthogonal basis can effectively isolate uncertainties caused by partial observability. Dynamic decomposition provides a structural foundation for building general-purpose world models.

\clearpage
\balance
\bibliographystyle{IEEEtran} 
\bibliography{aaai2026}

\end{document}